\documentclass[10pt,twocolumn,letterpaper]{article}
\usepackage{cvpr}
\usepackage{times}
\usepackage{epsfig}
\usepackage{graphicx}
\usepackage{amsmath}
\usepackage{amssymb}
\usepackage{booktabs}
\usepackage{color}
\usepackage{colortbl}
\usepackage{subfigure}

\usepackage[pagebackref=true,breaklinks=true,letterpaper=true,colorlinks,bookmarks=false]{hyperref}

\definecolor{cwblue1}{rgb}{0.27,0.427,0.623}
\definecolor{cwblue2}{rgb}{0.286,0.454,0.658}
\definecolor{cwblue3}{rgb}{0.733,0.811,0.905}

\newcommand{\para}[1]{\textbf{#1} ---}

\cvprfinalcopy % *** Uncomment this line for the final submission

\def\cvprPaperID{1683} % *** Enter the CVPR Paper ID here

\ifcvprfinal\fi

\def\h{{\mathbf{h}}}
\def\c{{\mathbf{c}}}
\def\U{{\mathbf{U}}}
\def\sigmoid{\sigma}

\begin{document}

%%%%%%%%% TITLE
\title{Learning to detect and localize many objects from few examples}

\author{Bastien Moysset\\
A2iA SAS,Paris, France\\
INSA-Lyon, LIRIS, UMR5205,F-69621\\
%{\tt\small bm@a2ia.com}
\and
Christopĥer Kermorvant\\
Teklia SAS, Paris, France\\
\and
Christian Wolf\\
Universit\'{e} de Lyon, CNRS, France\\
INSA-Lyon, LIRIS, UMR5205,F-69621\\
}

\maketitle
%\thispagestyle{empty}

%%%%%%%%% ABSTRACT
\begin{abstract}
\noindent
The current trend in object detection and localization is to learn predictions with high capacity deep neural networks trained on a very large amount of annotated data and using a high amount of processing power.
In this work, we propose a new neural model which directly predicts bounding box coordinates. The particularity of our contribution lies in the local computations of predictions with a new form of local parameter sharing which keeps the overall amount of trainable parameters low. Key components of the model are spatial 2D-LSTM recurrent layers which convey contextual information between the regions of the image. 

We show that this model is more powerful than the state of the art in applications where training data is not as abundant as in the classical configuration of natural images and Imagenet/Pascal VOC tasks.  We particularly target the detection of text in document images, but our method is not limited to this setting. The proposed model also facilitates the detection of many objects in a single image and can deal with inputs of variable sizes without resizing.
%Experiments have been performed on a document analysis task, the localization of the text lines in the Maurdor dataset.

\end{abstract}

\section{Introduction}
\noindent
Object detection and localization in images is currently dominated by approaches which first create proposals (hypothesis bounding boxes) followed by feature extraction and pooling on these boxes and classification, the latter steps being usually performed by deep networks \cite{girshick2014rich,GirshickFastRCNN2015,GirshickFasterRCNN2015,YOLOCVPR2016,Bell2016InsideOut}. Very recent methods also use deep networks for the proposal step \cite{GirshickFasterRCNN2015,YOLOCVPR2016,LiuErhanSSD2016}, sometimes sharing features between localization and classification. Differences exist in the detailed architectures in the way calculations are shared over layers, scales, spatial regions etc. (see section \ref{sec:objectives} for a detailed analysis). Another criterion is the coupling between hypothesis creation and confirmation/classification. Earlier works create thousands of hypotheses per image, sometimes using low level algorithms (e.g. R-CNN \cite{girshick2014rich}), leaving the burden of validation to a subsequent classifier. Current work tends to create very few proposals per image, which satisfy a high degree of ``objectness''.

\begin{figure}[t] \centering
\includegraphics[width=8cm]{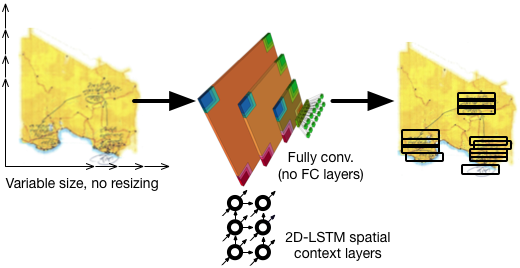}
\caption{\label{fig:teaser}A fully convolutional model with high spatial parameter sharing and fully trainable 2D-LSTM context layers learns to detect potentially many objects from few examples and inputs of variable sizes.}
\end{figure}

\begin{figure*}[t] \centering
\includegraphics[width=18cm]{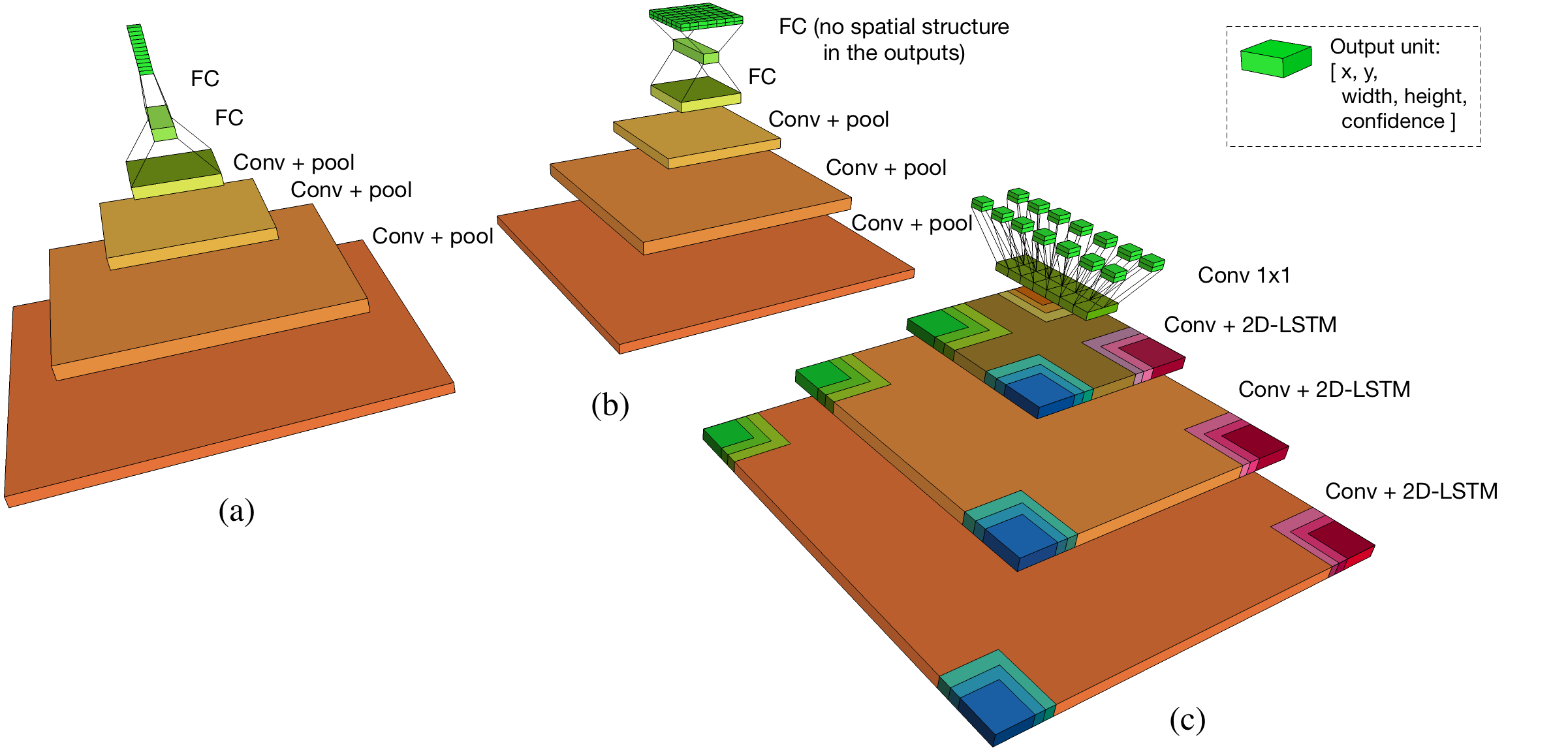}
\caption{\label{fig:yolovssdl}Sketches of different ways to model bounding box regression spatially (we do not show the correct numbers of layers and units). (a) Multibox \cite{erhan2014scalable}; (b) Yolo \cite{YOLOCVPR2016} and SSD \cite{LiuErhanSSD2016}; these three models pass through a fully connected layer, which connects to a set of bounding box outputs; architecture wise, these three models are identical. The grid-like display of the outputs in (b) is due to training; (c): our proposed model is fully-convolutional (no fully-connected (=FC) layers) and keeps spatial structure in the feature map. Global context is handled through 2D-recurrent LSTM networks, indicated through colored squares starting in each corner of each feature map (see also figure \ref{fig:2dlstm}).}
\end{figure*}

In this work we focus on the localization step, targeting cases where the existing methods tend to give weak results:
\begin{itemize}
\item the current trend is to design high capacity networks trained on large amounts of training data either directly or as a pre-training step. However, in some applications, the image content is only very weakly correlated to the data available in standard dataset like Imagenet. In the case of small and medium amounts of training data, fully automatic training of deep models remains a challenge in these cases.

\item we allow for the detection and localization of a relatively high number of potentially small objects in an image, which is especially hard for existing methods \cite{Bell2016InsideOut}. Our target application is the localization of text boxes, but our method is not restricted to this kind of setting. 
\end{itemize}

\noindent
Similar to recent work, the proposed method localizes bounding boxes by direct regression of (relative) coordinates. The main contribution we claim is a new model which performs spatially local computations, efficiently sharing parameters spatially. The main challenge in this case is to allow the model to collect features from local regions as well as globally pooled features in order to be able to efficiently model context.

Similar to models like YOLO \cite{YOLOCVPR2016} and Single-Shot Detector \cite{LiuErhanSSD2016}, our outputs are assigned to local regions of the image. However, in contrast to these methods, each output is trained to be able to predict objects in its support region, or outside. Before each gradient update step, we \emph{globally} match predictions and ground truth objects. Each output of our model directly sees only a limited region of the input image, which keeps the overall number of trainable parameters low. However, outputs get additional information from outside regions through context, which is collected using spatial 2D recurrent (LSTM) units. This spatial context layer proved to be a key component of our model.

We propose the following contributions:
\begin{itemize}
\item A new fully convolutional model for object detection using spatial 2D-LSTM layers for handling spatial context with an objective of high spatial parameter sharing.
\item The capability of predicting a large number of outputs, made possible by the combination of highly local output layers ($1\times 1$ convolutions) and preceding spatial LSTM layers.
\item The possibility of predicting outputs from input images of variable size without resizing the input.
\item An application to document analysis with experiments on the difficult and heterogeneous Maurdor dataset, which show that the model significantly outperforms the state of the art in objects detection.
\end{itemize}

% Christian: je ne suis pas sur qu'il faut mettre en avant cela.
%Having small networks enable to avoid the need for expensive GPUs on the customer side or the need to convey confidential information to a remote server. It also enables the algorithm to work on systems with less available memory and comuting power like smartphones.

%- Represent well this type of problem with a high number of small objects to detect, small number of training data available and a potential envy to do the object detection on a smartphone and/or to keep the processed information confidential.

\noindent
The paper is organized as follows: the next section briefly outlines related work. 
Section \ref{sec:objectives} discusses properties and trade-offs of deep models related to convolutions, poolings and subsampling, which will be related to our proposed model.
Based on these conclusions, a new model is introduced in section \ref{sec:ourmethod}.

\subsection{Related work}

\noindent
Earlier (pre-deep learning) work on object recognition proceeded through matching of local features \cite{LoweSIFT2001} or by decomposing objects into mixtures of parts and solving combinatorial problems \cite{felzenszwalb2010object}. Early work on deep learning first extended the sliding window approach to deep neural networks. To avoid testing a large number of positions and aspect ratios, R-CNN \cite{girshick2014rich} introduced the concept object proposals, created by separate methods, followed by convolutional networks to classify each proposal. The concept was improved as Fast R-CNN \cite{GirshickFastRCNN2015} and Faster R-CNN \cite{GirshickFasterRCNN2015}. 

Erhan et al. proposed Multibox \cite{erhan2014scalable,ErhanMultiboxFollowup2015}, which performs direct regression of bounding box locations instead of relying on object proposals. After each forward pass, network outputs are assigned to target ground-truth boxes through a matching algorithm. YOLO \cite{YOLOCVPR2016} and the Single-Shot Detector \cite{LiuErhanSSD2016} can be seen as variants of this concept, they will be discussed in more detail in section \ref{sec:objectives}.

Some recent work strives to detect and localize objects with pixel-wise precision, which somewhat blurs the boundaries between object detection and semantic segmentation \cite{PinheiroNIPS2015DeepMask,PinheiroLinECCV16sharpmask}. Methods which learn to segment without pixelwise ground truth have also been proposed \cite{Pinheiro_2015_CVPR}. Pixelwise segmentation is not needed in our application, where the segmentation step is performed in a latter stage jointly with recognition (recognition results will be given in the experimental section).

Context through spatial 2D-recurrent networks has been proposed as early as in \cite{Graves2DLSTM2009}. However, up to our knowledge, no method did use it for object localization. Similarly to our method, inside-Outside-Nets \cite{Bell2016InsideOut} contain 2D spatial context layers collecting information from 4 different directions. However, the hidden transitions of recurrent layers are set to identity,  whereas our model contains fully-fledged trainable 2D-LSTM layers. Moreover, localization is performed as ROI proposals with selective search, the deep model being used only for classification and bounding box correction, whereas we do not require a region proposal step. Our model directly performs bounding box regression. Other recent work uses 2D recurrent networks for semantic segmentation \cite{ByeonSemanticSegmentationLSTM2015}.

CNNs have been used before for text detection, for instance in \cite{zhang2016multi}, a Fully convolutional network (FCN) is used to classify each position of a salient map as text or non-text. In \cite{GuptaZissermannTextCVPR2016}, a YOLO-related method is proposed for the detection of text in natural images but only few objects are present in the images.

The problem of dataset sizes has been addressed before, with strategies reaching from external memories \cite{DeepMindExternal2016} and unsupervised learning, for instance by learning feature extraction from physics \cite{StewardErmonPhysics4Unsupervised2016}.

\begin{figure*}[t] \centering
\subfigure{\includegraphics[width=16cm]{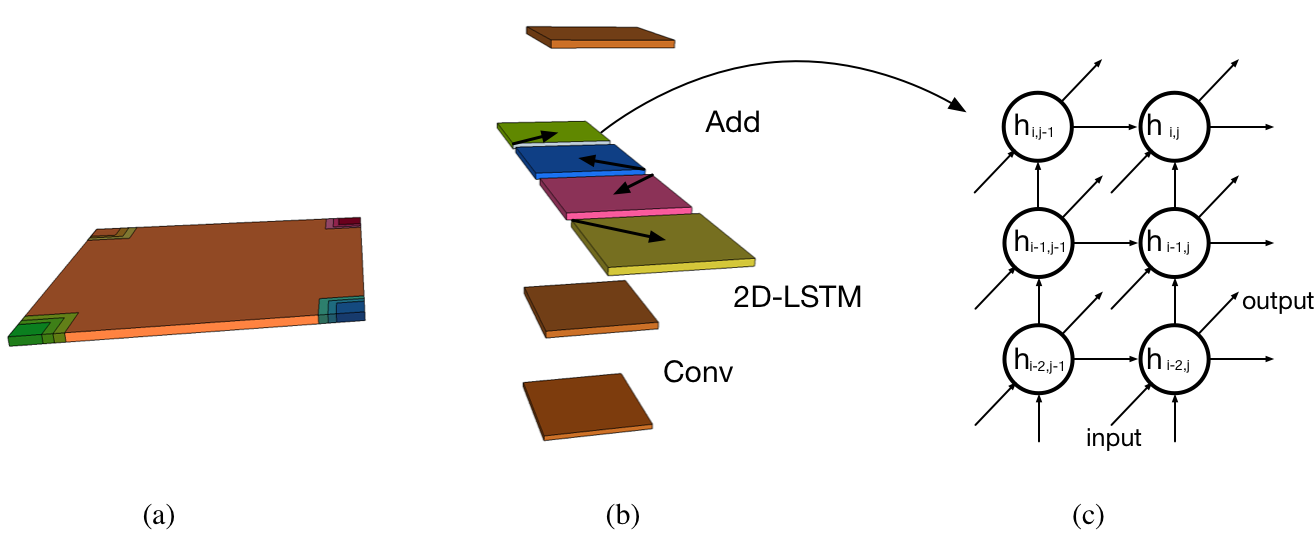}}
%\subfigure{\includegraphics[width=4cm]{figs/2d-lstm-onedir.png}}
\caption{\label{fig:2dlstm}(a) One of the conv+LSTM modules as shown in figure \ref{fig:yolovssdl}; (b) the module is composed of a convolutional layer and 4 directional 2D-LSTM layers in parallel, whose output feature maps are then element-wise added (not concatenated); (c) a single directional 2D-LSTM, shown for left-to-right/bottom-to-top direction. Each element gets recurrent connections from two different predecessors. Only a single unit is shown per site; RNN notation is used (memory cell/gates are not shown). In contrast to \cite{Bell2016InsideOut}, we use real LSTM models with trainable transition matrices.}
\end{figure*}

\section{Delving again into convolutions, pooling, strides and spatial structure}
\label{sec:objectives}

\noindent
Object detection and localization with convolutional deep neural networks is governed by a set of implicit properties and requirements, which we will try to lay out in the following lines. We will concentrate on the approach of direct prediction of object locations (as opposed to creating proposals from additional and not-tightly connected methods). The goal of this section is to discuss the effects and importances of each part and the trade-offs to consider in these architectures, which will lead us then to the formulation of the proposed model. 

The input image is passed through a series of convolutional layers, each of which extracts features from the preceding layer. Although not absolutely required, reducing the spatial size of the features maps (often combined with pooling) is frequently done in order to increase the receptive fields, i.e. the relative size of the filters w.r.t. to the inputs. Choosing when to pool and to reduce can be critical, and optimizations can lead to large decreases in the numbers of trainable parameters \cite{SqueezeNetOpenreview2017}. An alternative to in-between-layer pooling is changing the size of filters, especially as ``\emph{a trou}'' computation in order to keep the number of parameters low \cite{YuKoltunICLR2016ATrou}.

At some point, a model needs to collect features from a spatial support region. The way this pooling is distributed over the different layers will decide important properties of the model: 
\begin{itemize}
\item Classical networks stop the sequence of convolutions and reductions before the \emph{spatial} size of the feature map shrinks to $1{\times}1$, keeping a spatial / geometrical structure in the feature representation. Subsequent fully connected layers then perform further feature extraction, implicit pooling and decision taking.

The spatial structure of the feature map allows to perform controlled pooling from specific regions, but limits the shift-invariance of the representation.

\item More recently, fully-convolutional networks (FCN) perform convolutions and reductions+pooling until the spatial size of the feature map is negligible, e.g. $1{\times}1$, with a high feature dimension ($1{\times}1{\times}4096$ in the network for semantic segmentation proposed in \cite{long_shelhamer_fcn2015}). The goal here is to fully translate geometry and appearance into features and semantics. 

Training can in principle lead to a spatial structuring of the feature dimension, i.e. training can lead to a situation where different elements of the feature layer correspond to different regions in the input image. However, this is not  a constraint in the model and each activation of the last feature layer can potentially contain features from the full image.
\end{itemize}

\noindent
Object detection and localization require certain properties, like shift invariance, spatial precision and context collected from the global scene. Several state of the art models, Multibox \cite{erhan2014scalable}, YOLO \cite{YOLOCVPR2016} and Single-Shot Detector (SSD) \cite{LiuErhanSSD2016} tackle this through an architecture sketched in figure \ref{fig:yolovssdl}a and \ref{fig:yolovssdl}b\footnote{The purpose of this figure is to show the strategy these models use to translate geometry and resolution into features. In particular, we do not show the actual numbers of layers and units. For SSD\cite{LiuErhanSSD2016} , we do not show the way how this model handles multiple scales.}. A sequence of convolutions and reductions decrease the spatial size of feature maps down to a small grid ($7{\times}7$ for \cite{YOLOCVPR2016}, $9{\times}9$ for \cite{LiuErhanSSD2016}). This map is then fully connected to a $1{\times}1{\times}4096$ feature layer and again fully connected to a set of outputs, each output predicting bounding box positions and confidence scores (as well as class scores if required). This approach has several advantages. Each of the outputs is fully connected to previous layers and therefore potentially has access to information from the full image. The last feature layer mixes spatial structure and feature dimensions in a trainable way. 

Although there is no principled difference in how the last fully connected layer is actually implemented in the three models, we display the output layer differently for Multibox \cite{erhan2014scalable} (figure \ref{fig:yolovssdl}a) and for YOLO \cite{YOLOCVPR2016} or SSD \cite{LiuErhanSSD2016} (figure \ref{fig:yolovssdl}b). For the latter two, and also in accordance with the figures of the respective papers, the outputs are shown as 
a spatial grid ($7{\times}7$ for \cite{YOLOCVPR2016}, $9{\times}9$ for \cite{LiuErhanSSD2016}). However, this structuring is an interpretation, as the spatial structure of the grid is not wired into the network architecture. It is justified through the way training is performed in these models, in particular on the way  ground truth outputs are matched (assigned) to the network outputs. In the case of \cite{YOLOCVPR2016}, this assignment is purely spatial: outputs of a given cell are trained to provide predictions of a spatial region corresponding to this cell (see section \ref{sec:training}).

The main shortcoming of these models, which we will address in the next section, lies in the fully connected feature and output layers at the end. We argue that they limit invariance and contain too many parameters.

\section{A local spatially recurrent model}
\label{sec:ourmethod}

\noindent
We propose a new model designed to detect a large number of (potentially) small objects from a low number of training examples, i.e. with a model with a small number of trainable parameters. We achieve this with two techniques:

\textbf{A) Feature sharing} --- we predict different object locations from local features only. More precisely, the output layer of a single object bounding box is not fully connected to the previous layer, as illustrated in figure \ref{fig:yolovssdl}c. Outputs are connected through $1{\times}1$ convolutions, i.e. each element $(i,j)$ of the last feature map is connected to its own set of $K$ output modules, each module consisting of relative $x$ and $y$ positions, width, and height and a confidence score used to confirm the presence of an object at the predicted position. The objectives here are two-fold:

\begin{itemize}
\item To drastically reduce the number of parameters in the output layer by avoiding parameter hungry fully connected layers. 
%Indeed, the number of input neurons of the fully connected layer is divided by the size of the feature map as we use a vector of the feature map and not all of it. The number of output neurons of this fully connected layer can also be divided because we now want to be able to detect the maximum number of objects that can be present in a small part of the image and not the maximum number of objects that can be present in the whole image.

\item To share parameters between locations in the image, increasing shift invariance and significantly reducing the requirements for data augmentation.
\end{itemize}

\textbf{B) Spatial recurrent context layers} --- the drawback of local parameter sharing is twofold: i) objects may be larger than the receptive field of each output, and ii) we may lose valuable context information from the full input image. We address both these concerns through context layers consisting of Multi-Dimensional Long-Short term memory models \cite{Graves2008RnnHandwriting}, which are inserted between the convolutional layers. These MD-LSTM layers aim at recovering the context information from the area outside of the receptive field.

Figure \ref{fig:2dlstm} illustrates how the context layers are organized. Each convolutional layer is followed by 4 different parallel 2D-LSTM layers, which propagate information over the feature map elements in 4 different diagonal directions, starting at the 4 edges. For each of the directions, each element gets recurrent connections from 2 different neighbouring sites. The outputs of the 4 directions are summed ---  concatenation would have been another possibility, albeit with a drastically higher amount of parameters. No pooling is performed between the convolutions. Spatial resolution is reduced through convolutions with strides between 2 and 4 (see table \ref{tab:architecture}). 

The network outputs are computed from the last hidden layer as a regression of the \emph{normalized relative} bounding box locations. In particular, the absolute location of each predicted object is calculated by multiplying the network output with a width parameter vector $\Lambda$ and an offset vector $\Delta$, whose values depend on the architecture of the network. More formally, the location $l_{i,j,k}$ for the $k^{th}$ object prediction of element $(i,j)$ of the last feature map is given as follows:
\begin{equation}
\label{eq:outputs}
l_{i,j,k}= \Lambda^T \sigmoid\left( {\U_k} \h_{i,j} + \c_k \right)  + [ i{-}1 \ \ j{-}1 ]^T \Delta
\end{equation}
where $\h$ is the last hidden layer, $\sigmoid(\cdot)$ is the element-wise sigmoid function and the weights $\U_k$ and biases $c_k$ are trainable parameters. Note that, since the outputs are $1{\times}1$ convolutions, the parameters $\{\U_k,c_k\}$ are shared over locations $(i,j)$. However, each object predictor $k$ features its own set of parameters.

\textbf{Flexibilty} --- another significant advantage of the proposed local method is that we can handle images of varying sizes without performing any resizing or cropping. Decreasing or increasing the size of the input image, or changing its aspect ratio, will change the size of the post convolutional feature maps accordingly. This will change the number of network outputs, i.e. object predictions.

\section{Training}
\label{sec:training}

% Jamais deux titres qui se suivent
%\subsection{Our training}

\noindent
The model is trained with stochastic gradient descent (SGD) using mini-batches of size 8 and dropout for regularization.
During training, object predictors (network outputs) need to be matched to targets, i.e. to groundtruth object positions. Similar 
to the strategy in MultiBox \cite{erhan2014scalable}, this is done globally over the entire image, which allows each bounding box predictor to respond to any location in an image.

We denote by $M$ the number of predicted objects, given as $M=I*J*K$, with $I$ and $J$ being the width and the height of the last feature map and $K$ the number of predictors per feature map location; we denote by $N$ the number of reference objects in the image. Matching is a combinatorial problem over the matching matrix $X$, where $X_{nm}{=}1$ when hypothesis $m$ is matched to target $n$, and 0 otherwise. For each forward-backward pass for each image, $X$ is estimated minimizing the following cost function:
\begin{equation}
\begin{split}
\label{eq:globalCost}
%C= \sum_{n=0}^N \sum_{m=0}^M X_{nm} \left( \alpha \left\|l_{m}-t_{n}\right\|^2 - \log{\left(\frac{c_{m}}{1-c_{m}}\right)} \right) - \log(1-c_{m})
Cost = \sum_{n=0}^N \sum_{m=0}^M X_{nm} \left( \alpha \left\|l_{m}-t_{n}\right\|^2 - \log(c_{m})\right) \\
- (1 - X_{nm}) \log(1-c_{m}) 
\end{split}
\end{equation}
%We define for each $m \in \left[0,M\right]$, 
where $l_{m}$ is a vector of size 4 corresponding to a predicted location, $c_{m}$ is the corresponding confidence,
% We also define for each $n \in \left[0,N\right]$, 
and $t_{n}$ as a target location. The first term handles location alignment, the second favours high confidence and $\alpha$ is a weight between both terms.

Equation (\ref{eq:globalCost}) is minimized subject to constraints, namely that each target box is matched to at most one hypothesis box and vice versa. This is a well known bi-partite graph matching problem, which can be solved with the Hungarian algorithm \cite{munkres1957algorithms}. Equation (\ref{eq:globalCost}) gives the loss function used in the SGD parameter updates. However, we prefer to set different values of $\alpha$ for gradient updates and for matching. We found it important to increase $\alpha$ for the matching in order to help the network to use more diverse outputs.

%  We create a local cost $C_{nm}$ between each pair $n,m$ as described in equation \label{eq:localCost}. The global 
% cost is minimized using the Hungarian algorithm \cite{munkres1957algorithms} on these local costs to obtain the $X$ matching matrix. 
% The use of the Hungarian algorithm ensures that each target box is matched to an hypothesis box and that an hypothesis box is not matched to more than one 
% target box.
% Note that, if $\alpha_{match}$ has the same role as $\alpha_{cost}$, they do not have to be equal. In practice, we found that it is better to take $\alpha_{match}$ higher than $\alpha_{cost}$ 
% in order to help the network to use more different outputs.

% \begin{equation}
% \label{eq:localCost}
% C_{nm} = \alpha_{match} \left\|l_{m}-t_{n}\right\|^2 - \log\left(\frac{c_{i}}{1-c_{i}}\right)
% \end{equation}

%\noindent
%\para{With respect to related work}
As mentioned earlier, our matching strategy is similar to the one described in MultiBox \cite{erhan2014scalable} and has the same global property brought by the confidence term (albeit applied to local outputs, compared to the global outputs in \cite{erhan2014scalable}). On the other hand, in SSD\cite{LiuErhanSSD2016} and YOLO\cite{YOLOCVPR2016}, matching is done locally, i.e. predictors are matched to targets falling into spatial regions they are associated with. This is the reason for the spatial interpretation of the output grid shown in figure \ref{fig:yolovssdl}. 
% In particular, each predictor is matched to respectively, the four box 
% coordinates or the two coordinates of the center of the box. 
% The use of the Hungarian algorithm for the matching, as in SSD and MultiBox, enables to associate all our target boxes to hypothesis boxes. On the contrary, 
Moreover, YOLO matches only one target location with each spatial cell, which leads to non-matched targets in the case of several objects with bounding box centers in the same cell. In our target application, where a large number of objects may be present, a large number of objects will not be matched to any predictor during training, as can be seen in the example in figure \ref{fig:yoloProblem}.

In SSD and MultiBox, the matching process is restricted to a fixed dictionary of anchor locations obtained arbitrarily\cite{LiuErhanSSD2016} or with clustering\cite{erhan2014scalable}, which helps the network to create outputs specialized to regions in the image. This was proved unnecessary and even counter-productive in our case, where predictors share parameters spatially.

\section{Experimental results}

% Jamais deux titres qui se suivent
%\subsection{Training data}
\noindent
We tested the proposed model and the baselines on the publicly available Maurdor dataset \cite{Brunessaux2014}. This highly heterogeneous dataset is composed of 8773 document images ( train:6592; valid:1110; test:1071 ) in mixed French, English and Arabic text, both handwritten and printed.

The dataset is annotated at paragraph level. For this reason, we use the technique detailed in \cite{bluche2014automatic} to get annotation at line level and we keep only the pages where we are confident that the automatic line position generation has worked well. We obtain a restricted dataset containing 3995 training pages, 697 validation pages and 616 test pages that are used for training, validation and test for the evaluation of intersection over union and detEval metrics.

For the Bag of Word metric, we evaluate on the 265 pages fully in English and on the 507 pages fully in French of the full Maurdor test set in order to avoid the line language classification task.

\subsection{Metrics}
\noindent
We evaluate the performance of our method using three different metrics:

\para{Intersection over union}
IoU is a commonly used metric in object detection and image segmentation. It is given as the ratio of the intersection and the union of reference  and hypothesis objects. Reference objects and hypothesis objects are matched by thresholding their IoU score. In most frequent versions, only one hypothesis can be associated to a reference box, the others are considered as error/insertions. Alternatively to reporting IoU directly, after thresholding IoU, an F-Measure can be computed from Precision and Recall.

\para{DetEval}
DetEval \cite{WolfIJDAR2006} is the metric chosen for the ICDAR robust reading series of competitions. Its main advantage is that it allows many-to-many matchings between reference objects and hypothesis objects, which is important in applications where fragmentation of objects should be allowed (and eventually slightly punished), which is the case in text localization. Objects are assigned by thresholding overlap, and Precisions, Recall and F-Mesure are reported.

\para{Bag-of-words recognition error}
BoW is a goal oriented method described in \cite{pletschacher2015europeana}, which measures the performance of a subsequent text recognition module. The objective is to avoid the need of judging the geometrical precision of the result and to directly evaluate the performance of the goal of any localization method. In the case of the target application this is the subsequent recognizer.

We use the recognition model from \cite{A2IARecognizer2014}, which is based on deep convolutional networks and spatial/sequence modelling with 2D-LSTM layers. Assigning character labels to network outputs is performed with the Connectionist Temporal Classification framework \cite{CTC_ICML2006}. Recent follow-up work solves this problem with attention based mechanisms \cite{BlucheNIPS2016}, this will be investigated in future work. The recognizer is trained on both handwritten and printed text lines, separately on English and French text. We apply them on crops of localized bounding boxes. The Bag of Word metric, on the contrary to metrics based on the Levenshtein distance enables to avoid an alignment that can be ambiguous at page level. Word insertion and deletions are computed at page level and F-Measure is reported.

\subsection{Baselines}

\noindent
\para{Traditional text segmentation methods}
For comparison, we used two techniques based on image processing (w/o machine learning) for document text line segmentation. Shi et al. \cite{Shi2009a} use steerable directional filters to create an adaptive local connectivity map. Line locations are given by the positions of the connected components extracted from the binarisation of this connectivity map. These positions are refined with heuristic-based post-processing. The method proposed by Nicolaou et al. \cite{Nicolaou2009} follows the whitest and blackest paths in a blurred image to find lines and interlines. 

\noindent
\para{Yolo and MultiBox}
For YOLO, we used two classes for the object classification part of the model, handwritten text lines and printed text lines. It helped the model to learn better than without classification.

Both systems were tested in two different configurations: the original architecture tuned for large scale image recognition, and an architecture which we optimized for our task on the validation set. In particular, the size of the filters was adapted to the shape of the objects. Hyper-parameter tuning led in both cases to architectures with heavily reduced numbers of layers and less units per layer. We also optimized learning rates and minibatch sizes.

\subsection{Architectures}

\noindent
The network architecture of the proposed model has been tuned to correspond to our task. The found hyperparameters are detailed in Table \ref{tab:architecture}. The width and height of the feature maps is given for illustration but it can of course vary. In particular, the aspect ratio of the image can vary. We would like to stress again that the number of parameters is independent of the actual size of the input image. 

The inputs of our network are raw gray-scaled images with width normalisation. The use of color images was not improving the results on our task.

Note that the number of weights in our last layer, the position prediction layer, is rather small : 3,700. To be able to predict the same number of objects, with the same number of input features, MultiBox\cite{erhan2014scalable} and Yolo\cite{YOLOCVPR2016} would have needed 15,688,200 parameters.

\def\NA{} % previously {/}
\begin{table}
\begin{center}
\caption{Network architecture/hyper-parameters. The input and feature map sizes are an illustrative example. The number of parametres does \emph{NOT} depend on the size of the input image.}
\label{tab:architecture}
\begin{tabular}{lcccr}
Layer & Filter  & Stride & Size of the  & Number of  \\ %& feature map size??
      & size    &        & feature maps & parameters \\
\arrayrulecolor{cwblue1} \toprule  
Input & / & / & 1$\times$(598$\times$838) & \NA \\
C1  & 4$\times$4 & 3$\times$3 & 12$\times$(199$\times$279) & 204 \\
LSTM1 & / & / & " " & 8880 \\
C2 & 4$\times$3 & 3$\times$2 & 16$\times$(66$\times$139) & 2320 \\
LSTM2 & / & / & " " & 15680 \\
C3 & 6$\times$3 & 4$\times$2 & 24$\times$(16$\times$69) & 6936 \\
LSTM3 & / & / & " " & 35040 \\
C4 & 4$\times$3 & 3$\times$2 & 30$\times$(5$\times$34) & 8670 \\
LSTM4 & / & / & " " & 54600 \\
C5 & 3$\times$2 & 2$\times$1 & 36$\times$(2$\times$33) & 6516 \\
Output & 1$\times$1 & 1$\times$1 & 5$\times$20$\times$(2$\times$33) & 3700 \\
\end{tabular}
\end{center}
\end{table}

For training, we used a learning rate of $10^{-4}$ and mini-batches of size 8. Dropout with 0.5 probability is applied after each 2D-LSTM layer. The $\alpha$ parameter in equation (\ref{eq:globalCost}) is set to 1000 for matching and to 100 for weight updates during SGD.

We experimentaly found that resolution reduction between layers works better using strides $>1$ of the convolutional layers instead of max-pooling. This can be explained by our need for precision, while max-pooling is known to lead to shift invariance.

\subsection{Results and discussion}

\noindent
\textbf{Localization} results on the restricted Maurdor test set, for our proposed method and baselines, are shown with the IoU metric in Table \ref{tab:resultsiou} and with the DetEval metric in Table \ref{tab:resultsdeteval}. \textbf{Text recognition results} (on text objects localized by our method) are shown in table \ref{tab:resultsbow}, respectively for pages fully in French and fully in English of the whole Maurdor test set.

\begin{figure}[t] \centering
\includegraphics[width=6cm]{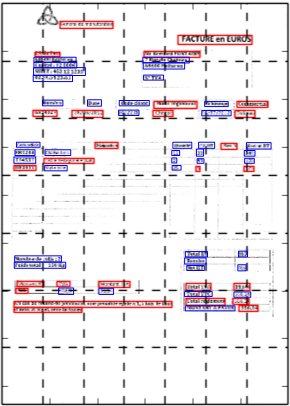}
\caption{\label{fig:yoloProblem}YOLO matches target objects according to the spatial region in which they fall in the image, which leaves many targets unassigned (shown in blue). In applications with many objects, this makes the method non applicable.}
\end{figure}

\begin{table}[!t]
\begin{center}
\caption{Detection performance: F-Measure with various thresholds(T) on IoU. On the restricted Maurdor test set (616 pages).}
\label{tab:resultsiou}
\begin{tabular}{lrrr}
Method & \multicolumn{3}{c}{------ F-Measure ------} \\
 & T=0.3 & T=0.5 & T=0.7\\
\arrayrulecolor{cwblue1} \toprule  
Shi et al. \cite{Shi2009a} & 40.7\% & 31.1\% & 21.1\% \\
Nicolaou et al. \cite{Nicolaou2009} & 36.1\% & 26.3\% & 15.9\% \\
\arrayrulecolor{cwblue1} \toprule  
Multibox \cite{erhan2014scalable} & 11.3\% & 2.1\% & 0.2\% \\
Multibox \cite{erhan2014scalable} (optimized) & 48.7\% & 23.0\% & 5.2\% \\
% YOLO \cite{YOLOCVPR2016} & \multicolumn{3}{c}{[not applicable --- cf. fig. \ref{fig:yoloProblem}]} \\
% YOLO \cite{YOLOCVPR2016} (optimized) & \multicolumn{3}{c}{[not applicable --- cf. fig. \ref{fig:yoloProblem}]} \\
\arrayrulecolor{cwblue1} \toprule  
Ours, no LSTMs & 49.9\% & 23.7\% & 5.3\% \\
Ours & 73.8\% & 43.6\% & 14.1\% \\
\end{tabular}
\end{center}
\end{table}

\begin{table}[!t]
\begin{center}
\caption{Detection performance with detEval\cite{WolfIJDAR2006}. On the restricted Maurdor test set (616 pages).}
\label{tab:resultsdeteval}
\begin{tabular}{lrrr}
Method & Recall & Precision & F-Meas.\\
\arrayrulecolor{cwblue1} \toprule  
Shi et al. \cite{Shi2009a} & 35.1\% & 38.4\% & 36.7\% \\
Nicolaou et al. \cite{Nicolaou2009} & 46.7\% & 39.6\% & 42.9\% \\
\arrayrulecolor{cwblue1} \toprule  
Multibox \cite{erhan2014scalable} & 4.2\% & 10.0\% & 6.0\% \\
Multibox \cite{erhan2014scalable} (optimized) & 28.8\% & 52.3\% & 31.1\% \\
% YOLO \cite{YOLOCVPR2016} & \multicolumn{3}{c}{[not applicable --- cf. fig. \ref{fig:yoloProblem}]} \\
% YOLO \cite{YOLOCVPR2016} (optimized) & \multicolumn{3}{c}{[not applicable --- cf. fig. \ref{fig:yoloProblem}]} \\
\arrayrulecolor{cwblue1} \toprule  
Ours, no LSTMs  & 28.6\% & 52.4\% & 31.1\% \\
Ours & 51.2\% & 61.4\% & 55.9\% \\
\end{tabular}
\end{center}
\end{table}

\begin{table}[!t]
\begin{center}
\caption{Detection and recognition performance: word-recognition F-Measure in BOW mode on the full English or French Maurdor test set.}
\label{tab:resultsbow}
\begin{tabular}{lrr}
Method & \begin{tabular}{@{}c@{}}French \\ (507 pages)\end{tabular} & \begin{tabular}{@{}c@{}}English \\ (265 pages)\end{tabular} \\
\arrayrulecolor{cwblue1} \toprule  
Shi et al. \cite{Shi2009a} & 48.6\% & 30.4\% \\
Nicolaou et al. \cite{Nicolaou2009} & 65.3 \% & 50.0\% \\
\arrayrulecolor{cwblue1} \toprule  
Multibox \cite{erhan2014scalable} & 27.2\% & 14.8\% \\
% Multibox semi-tuned \cite{erhan2014scalable} & 29.0\% & 30.0\% \\
Multibox \cite{erhan2014scalable} (optimized) & 32.4\% & 36.2\% \\
% YOLO \cite{YOLOCVPR2016} & \multicolumn{2}{c}{[not applicable --- cf. fig. \ref{fig:yoloProblem}]}\\
% YOLO \cite{YOLOCVPR2016} (optimized) & \multicolumn{2}{c}{[not applicable --- cf. fig. \ref{fig:yoloProblem}]} \\
\arrayrulecolor{cwblue1} \toprule  
Ours, no LSTMs & 57.8\% & 56.9\% \\
Ours & 71.2\% & 71.1\% \\
\end{tabular}
\end{center}
\end{table}

\begin{figure*}[t]
   \centering
   \subfigure[]{\includegraphics[width=2.7cm]{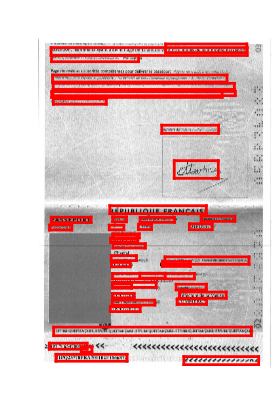}}
   \subfigure[]{\includegraphics[width=2.7cm]{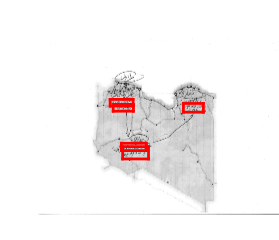}}
   \subfigure[]{\includegraphics[width=2.7cm]{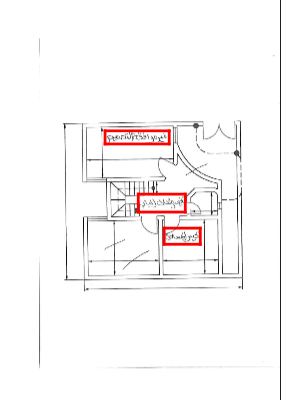}}
   \subfigure[]{\includegraphics[width=2.7cm]{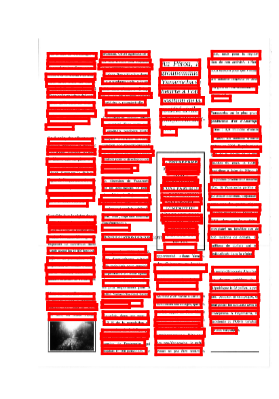}}
   \subfigure[]{\includegraphics[width=2.7cm]{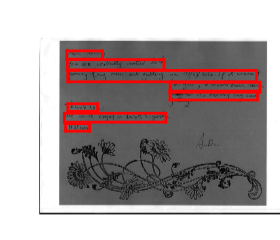}}
   \subfigure[]{\includegraphics[width=2.7cm]{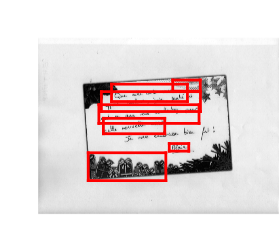}}
   \caption{Samples of results obtained with the proposed method on images of the Maurdor test set. The actual inputs are shown.}
   \label{fig:illustrationsResults}
\end{figure*}

For the IoU metric, results are given in Table \ref{tab:resultsiou}. We report F-Measure for different thresholds on IoU, i.e. for different localization quality requirements. The image-based techniques Shi et al. \cite{Shi2009a} and Nicolaou et al. \cite{Nicolaou2009} perform poorly when the threshold is low, i.e. when we are interested in the ability of the system to detect all the boxes regardless of the exact location. They suffer of low general recall. However, they are relatively precise. F-Measure drops less than the learning based methods when the precision requirements are increased by increasing the threshold on IoU. This can be explained by the nature of these algorithms, which proceed by binarisation of the input images. In the normal operating range of these algorithm, when the segmentation steps work out well, precision is almost guaranteed to be high. However, once images don't fall into the situations the algorithms has been tuned for, performance breaks down.

On the other hand, methods based on direct regression as MultiBox \cite{erhan2014scalable} and our proposed method are more robust and achieve better general recall, an advantage which is bought with a slight drop in precision. Our proposed method gives the best results for  realistic thresholds. The detEval metric results shown in Table \ref{tab:resultsdeteval} confirm the results from the intersection over union metric.

From the application perspective, namely the full page text recognition in documents, Table \ref{tab:resultsbow} shows that the proposed method delivers good results with over 70\% F-Measure Bag of Word score on both French and English, outperforming all other methods. This can be explained by its high recall, while the slightly better precision of image based methods is not an advantage since a recognizer can compensate for it, up to a certain limit.

The last two lines of Tables \ref{tab:resultsiou}, \ref{tab:resultsdeteval} and \ref{tab:resultsbow} illustrate the importance of adding 2D-LSTM layers to recover information as it significantly improves performances for all the metrics. The power of the 2D-LSTM layers can also be shown in Figure \ref{fig:illustrationsResults}, which gives some example detections. Figures \ref{fig:illustrationsResults}a and \ref{fig:illustrationsResults}b show that the model is capable of detecting objects which are larger than the receptive fields of the individual bounding box predictors. This is made possible through the context information gathered by the LSTM layers. Figures \ref{fig:illustrationsResults}a and \ref{fig:illustrationsResults}d show that the system is capable of detecting and locating a large number of small objects. 

Multibox \cite{erhan2014scalable} is significantly outperformed by our method, even if we optimize its hyper-parameters (on the validation set). We attribute this to the fact, that the output layers are not shared. The model needs to express similar prediction behavior for each output, thus relearn the same strategies several times. 

YOLO \cite{YOLOCVPR2016} proved to be impossible to apply to this kind of problem, at least in its current shape. As reported by the authors, the system gives excellent results on the tasks it has been designed for. However, despite extensive tuning of its hyperparameters, we were not able to reach satisfying results, although we worked with two different implementations: the original implementation of the authors, as well as our own implementation. We did identify the problem, however. YOLO has been designed for a small number of objects, with a predictor/target matching algorithm adapted to these settings (see also section \ref{sec:training}). As mentioned, only one target can be associated to each spatial cell, which is a harmless restriction for traditional object detection tasks. However, this is a real problem in our case, as shown in the example image in Figure \ref{fig:yoloProblem}. A large part of the ground truth objects in most figures will not be assigned to any predictor, and not trained for. Not only are these boxes missing at training, network outputs predicting their locations will be punished at the next parameter update, further hurting performance and hindering the networks from converging properly.

\subsection{Implementation}

\noindent
No deep learning framework was used for the implementation of the proposed method, since, until recently and the Theano version from Doetsch et al. \cite{doetsch2016returnn}, no 2D-LSTMs implementation was, up to our knowledge, yet existing in Tensorflow, Torch, Theano or Caffe. The system has been implemented using our inhouse framework implemented in C++, including the SG optimizer, dropout etc. For this reason also, the model has been trained on CPUs.

For YOLO we used two different implementations. We implemented and trained our own implementation in Tensorflow, and we also used the official source code published by the authors \footnote{\url{http://pjreddie.com/darknet/yolo}}.

\section{Conclusion}

\noindent
We presented a new fully-convolutional model for the detection and localization of a potentially large number of objects in images. To optimize invariance and in order to limit the number of trainable parameters, we shared parameters of the output layer over spatial blocks of the image, implementing the output layer as $1{\times}1$ convolution. To deal with objects which are larger than the receptive field, and in order to allow the model to collect features from the global context, we added 2D-LSTM layers between the convolutional layers.

We compared the proposed model to the state of the art in object detection, in particular to YOLO \cite{YOLOCVPR2016} and Multibox \cite{erhan2014scalable}. We measured detection performance and word recognition performance of a subsequent classifier. Our experiments showed, that the proposed model significantly outperforms both methods, even if their hyper-parameters are optimized for the targeted configurations.

{\small
\bibliographystyle{ieee}
\bibliography{refs}
}

% \newpage
% \input{sec_suppl_material}

\end{document}